\documentclass[manuscript]{acmart}
\AtBeginDocument{%
  \providecommand\BibTeX{{%
    \normalfont B\kern-0.5em{\scshape i\kern-0.25em b}\kern-0.8em\TeX}}}

\begin{document}



\title{Exploring Object Status Recognition for Recipe Progress Tracking in Non-Visual Cooking}



\author{Franklin Mingzhe Li}
\affiliation{%
  \institution{Carnegie Mellon University}
  \city{Pittsburgh}
  \state{PA}
  \country{United States}
}
\email{mingzhe2@cs.cmu.edu}

\author{Kaitlyn Ng}
\affiliation{%
  \institution{Carnegie Mellon University}
  \city{Pittsburgh}
  \state{PA}
  \country{United States}
}
\email{kgn@andrew.cmu.edu}

\author{Bin Zhu}
\affiliation{%
  \institution{Singapore Management University}
  \city{Singapore}
  \country{Singapore}
}
\email{binzhu@smu.edu.sg}

\author{Patrick Carrington}
\affiliation{%
  \institution{Carnegie Mellon University}
  \city{Pittsburgh}
  \state{PA}
  \country{United States}
}
\email{pcarrington@cmu.edu}

\renewcommand{\shortauthors}{Li et al.}



\begin{abstract}
Cooking plays a vital role in everyday independence and well-being, yet remains challenging for people with vision impairments due to limited support for tracking progress and receiving contextual feedback. Object status — the condition or transformation of ingredients and tools — offers a promising but underexplored foundation for context-aware cooking support. In this paper, we present OSCAR (Object Status Context Awareness for Recipes), a technical pipeline that explores the use of object status recognition to enable recipe progress tracking in non-visual cooking. OSCAR integrates recipe parsing, object status extraction, visual alignment with cooking steps, and time-causal modeling to support real-time step tracking. We evaluate OSCAR on 173 instructional videos and a real-world dataset of 12 non-visual cooking sessions recorded by BLV individuals in their homes. Our results show that object status consistently improves step prediction accuracy across vision-language models, and reveal key factors that impact performance in real-world conditions, such as implicit tasks, camera placement, and lighting. We contribute the pipeline of context-aware recipe progress tracking, an annotated real-world non-visual cooking dataset, and design insights to guide future context-aware assistive cooking systems.

\end{abstract}




\maketitle

\section{Introduction}

Following recipes while cooking is an important yet challenging task for people with vision impairments, particularly due to limited support for tracking progress and receiving contextual feedback \cite{li2024recipe}. Prior research has shown that blind and low vision (BLV) individuals often rely on embodied strategies—such as spatial memory, tactile exploration, and environmental routines—to navigate the kitchen without vision \cite{li2021non,li2024recipe,wang2023practices,huh2025vid2coach}. These strategies support independence but are often brittle in unfamiliar or complex recipes, where confirming task completion, anticipating next steps, and recovering from disruptions are critical for both usability and safety \cite{li2024contextual, kostyra2017food, jones2019analysis}.

While tools like screen readers and smart speakers (e.g., Alexa, Google Home) enable audio-based access to recipes, they operate linearly and lack awareness of what is happening in the physical cooking space \cite{branham2019reading, abdolrahmani2018siri}. These systems cannot verify whether a step has been completed, adapt to changes in task flow, or reason about what remains to be done. As a result, many BLV users report uncertainty about their current progress, whether they have missed a step, or when it is safe to move on—especially when cooking alone or without pre-memorized routines \cite{li2024recipe,li2021non}. Prior work has emphasized the need for cooking technologies that provide context-aware feedback, rather than just access to static instructions \cite{li2024contextual}.

In parallel, computer vision and assistive technologies have enabled object and environment recognition to support non-visual interaction \cite{bigham2010vizwiz, guo2016vizlens, kim2022vision}. However, most systems focus on object presence, not object status — the condition or transformation of ingredients and tools. In cooking, this status is constantly changing: onions get chopped, sauces thicken, meat browns. Recent research shows that such transformations offer rich signals for tracking procedural progress \cite{xue2024learning}, but their potential to support accessibility remains underexplored. To investigate this gap, we ask:

\begin{itemize}
    \item RQ1: How useful is object status information in improving the performance of recipe step prediction using vision-language models?
    \item RQ2: How can object status information support reasoning about the current context of a cooking task, especially in real-world non-visual cooking scenarios?
\end{itemize}

To address these questions, we developed OSCAR (Object Status Context Awareness for Recipes), a technical pipeline (rather than a deployed assistive system) that explores the use of object status recognition to support recipe progress tracking in non-visual cooking. OSCAR integrates recipe parsing, object status extraction, visual alignment with cooking steps using vision-language models (VLMs) \cite{radford2021learning, zhai2023sigmoid}, and a time-causal model for structured prediction. Unlike tools that rely solely on recipe text or voice interaction, OSCAR reasons about the real-time cooking context through the evolving visual state of ingredients and tools — enabling both progress tracking and contextual feedback.

We emphasize that our focus is on evaluating the feasibility and performance of this technical pipeline through controlled dataset-based analysis, rather than assessing a complete assistive system in user-facing scenarios. We evaluate OSCAR across two datasets: (1) 173 instructional cooking videos from the YouCook2 dataset \cite{ZhXuCoAAAI18}, and (2) a real-world dataset of 12 non-visual cooking sessions recorded by BLV individuals in their homes. Our results show that object status improves step prediction accuracy by over 20\% across multiple VLMs. We also identify key factors affecting performance in real-world conditions, including implicit tasks, lighting, occlusion, and camera placement. Our work contributes:
\begin{itemize}
    \item A technical pipeline (OSCAR) for object status-driven recipe progress tracking, combining recipe modeling, vision-language alignment, and time-causal prediction.
    \item A real-world annotated dataset of 12 cooking sessions by people with vision impairments, reflecting diverse non-visual practices and challenges, which we will publicly release upon publication to support future accessibility and AI research.
    \item A technical evaluation demonstrating the feasibility and limitations of object status recognition for progress tracking in both instructional and real-world cooking contexts.
    \item An analysis of real-world performance factors and design implications for future context-aware assistive cooking technologies.
\end{itemize}

\section{Related Work}

We review related research in the areas of kitchen technologies and cooking support in HCI, followed by cooking practices and assistive tools for people with vision impairments. We then describe prior work on recipe modeling and vision-based progress tracking, and accessible recipe access. Finally, we situate our dataset contribution within disability-centered AI research, as well as dataset creation and evaluation practices.

\subsection{Cooking-related Technology and Accessibility Challenges}

Cooking is a complex, embodied task that demands real-time feedback, spatial awareness, and multi-step reasoning. In HCI and AI research, significant efforts have focused on enhancing cooking experiences through sensing, automation, and multimodal interaction. For instance, researchers have leveraged cameras and sensors to recognize object usage patterns and cooking actions \cite{olivier2009ambient,lei2012fine}. Lei et al. employed RGB-D cameras to detect fine-grained object interactions and cooking steps \cite{lei2012fine}, while others have explored smart appliance control and interactive assistance via voice, gesture, or mobile interfaces \cite{kim2017study,vu2018application}. Sensor-augmented tools like Lab-on-Spoon also demonstrate potential for real-time food monitoring by measuring temperature, color, and pH to ensure food safety and quality \cite{konig2015lab}.

However, most of these technologies are not designed with accessibility in mind and often overlook the unique challenges faced by people with vision impairments. Cooking for individuals who are blind or low vision (BLV) introduces distinct barriers due to the task's dynamic and visual nature. Prior studies document difficulties such as locating tools, assessing food states (e.g., browning, doneness), tracking task progress, and navigating cluttered or unfamiliar kitchens \cite{bilyk2009food,jones2019analysis,kostyra2017food}. Jones et al. found a strong correlation between vision loss severity and reduced cooking confidence and nutrition outcomes \cite{jones2019analysis}. As a result, many BLV individuals shift toward convenience foods or avoid complex recipes altogether.

To navigate these challenges, BLV cooks often develop compensatory strategies involving tactile organization, spatial memorization, or auditory cues like sizzling sounds \cite{li2021non}. Yet, these methods can be brittle—breaking down in unfamiliar environments or with complex, multistep dishes. Recent work by Li et al. highlights persistent issues in progress tracking, confirming completed actions, and determining next steps in the absence of visual cues \cite{li2021non, li2024recipe}. These findings underscore the need for accessible technologies that go beyond static recipe instructions, offering real-time, context-sensitive support tailored to the procedural and situational needs of non-visual cooking.

\subsection{Recipe Technologies and Non-Visual Progress Tracking}
Structured recipes are central to guiding cooking tasks and have been widely digitized for applications ranging from personalized meal planning to AI-driven cooking support. Prior research has explored recipe parsing, search, and recommendation through online platforms, voice interfaces, and annotated datasets \cite{Allrecip74:online, RecipesD31:online, tasse2008sour, marin2021recipe1m+, teng2011reciperecommendation,li2025oscar}. More recent efforts have focused on aligning recipe steps with instructional videos to track procedural content, enabling systems to provide step-by-step prompts or generate timelines for cooking actions \cite{lin2020recipe, cao2019video, gong2021temporal, donatelli2021aligning}.

Yet most of these systems are built under idealized assumptions—clear visual data, fixed action-object mappings, and known recipes. In real-world cooking, ingredients and tools often transform over time: raw ingredients become chopped, cooked, or mixed, making it difficult for vision-based systems to track progress through static labels alone. Xue et al. argue that recognizing object status changes (e.g., from raw to chopped to sautéed) offers a more accurate reflection of procedural progress than traditional object-verb detection \cite{xue2024learning}. Despite this insight, few systems explicitly model these transformations, and even fewer consider how they might support non-visual or assistive applications.

For people with vision impairments, recipe access typically relies on screen readers, smart speakers, or Braille displays, which translate static text into audio \cite{rodrigues2015getting, niran2016access, blenkhorn1995producing, shimomura2010accessibility}. While tools like Google Home and Amazon Alexa support basic recipe navigation through commands like “next step,” they lack awareness of user progress or confirmation of completed actions \cite{abdolrahmani2018siri, branham2019reading}. This limitation becomes problematic when users lose track of their place, face interruptions, or attempt to manage complex, multi-step processes without visual feedback.

Recent work has shown that BLV users often rely on internal strategies such as mental models, tactile routines, and step memorization to stay oriented during cooking \cite{li2024recipe}. However, these approaches are vulnerable to breakdowns, especially when switching tools, transitioning between stages, or reorienting after a distraction. While static access to recipe content is increasingly supported, real-time, context-sensitive feedback that reflects the procedural nature of cooking remains underexplored. Bridging structured recipe technologies with non-visual progress tracking presents a promising opportunity for more inclusive and supportive cooking systems.

\subsection{Dataset for Accessibility Research}

Recent advances in accessibility research have increasingly embraced the creation and use of specialized datasets to evaluate technical feasibility, benchmark system performance, and train models tailored to the needs of people with disabilities \cite{gurari2019vizwiz}. These datasets are not only vital for developing robust machine-learning systems but also serve as a foundation for understanding the everyday contexts, preferences, and constraints of disabled users.

One influential example is the ORBIT dataset, which provides a real-world benchmark for few-shot object recognition. It was constructed by blind users who captured their own personal items across varied home settings, resulting in a dataset that reflects authentic use cases and user environments \cite{massiceti2021orbit}. Other work, such as that by Theodorou et al., has called for a disability-first approach to dataset design, emphasizing the importance of representing the practices, tools, and goals of disabled individuals rather than retrofitting general-purpose datasets to accessibility tasks \cite{theodorou2021disability}.

Additional datasets have targeted domains such as assistive navigation, scene description, and input technologies. For instance, SANPO captures real-world pedestrian navigation tasks with visual impairments \cite{waghmare2025sanpo}, while VizWiz dataset focuses on visual question answering using images taken by blind users in natural environments \cite{gurari2019vizwiz}. Others have explored datasets for technology-mediated object search, text entry, and gesture-based interaction \cite{findlater2020input}, reinforcing the need for multimodal, real-world data that reflects disabled users’ information needs and interaction contexts.

Collectively, these efforts illustrate a growing recognition that dataset creation is a critical site of access work—not just a technical step, but a design choice that shapes what problems are solvable and for whom. While existing datasets often center on static scenes or discrete recognition tasks, our work builds on this disability-centered trajectory by capturing real-world, user-generated cooking practices that reflect the procedural, tactile, and adaptive nature of non-visual cooking. By documenting these practices and aligning them with structured recipe steps, we contribute a novel resource for evaluating accessible AI systems in the kitchen—one grounded in authentic behaviors rather than idealized laboratory conditions.

\begin{figure*}[t]
    \centering
    \includegraphics[width=1\columnwidth]{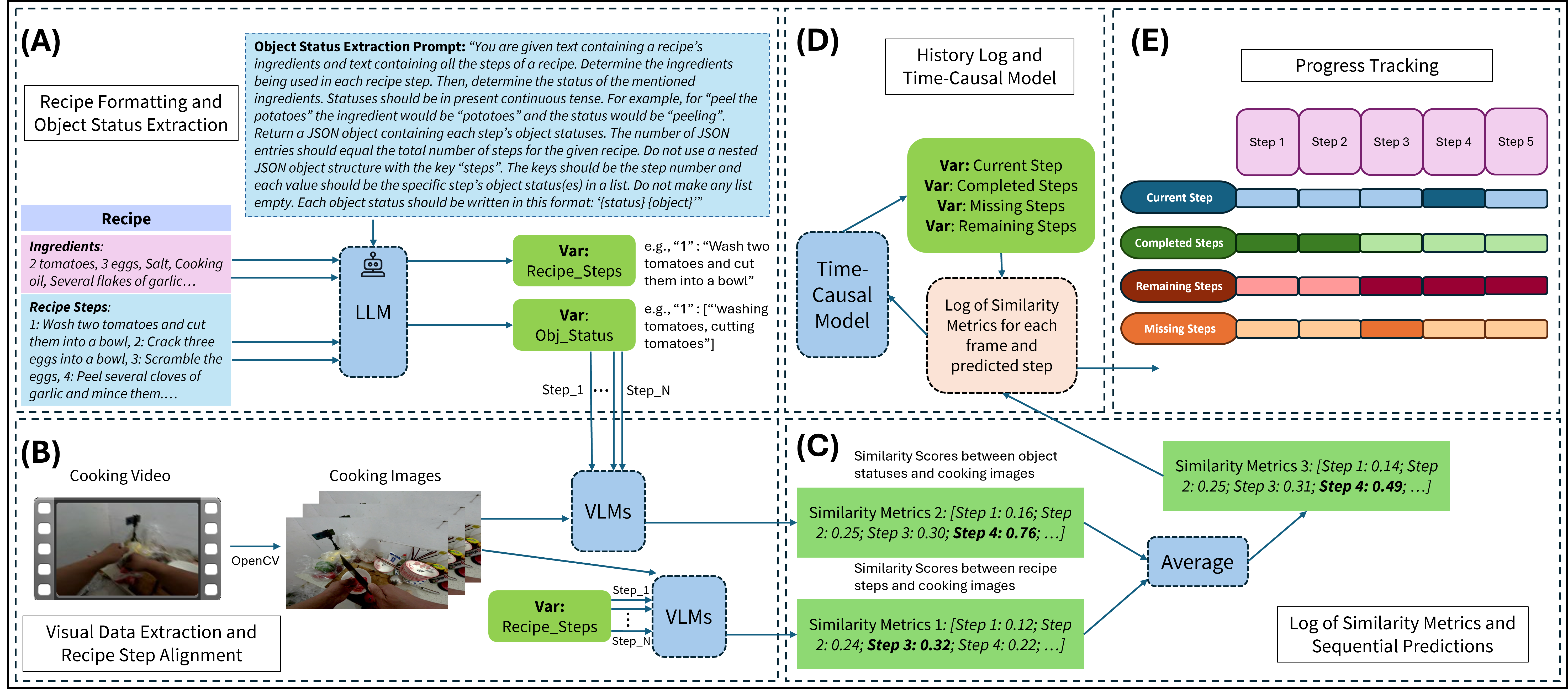}
    \caption{Illustration of OSCAR. (A) Recipe Formatting and Object Status Extraction, (B) Visual Data Extraction and Recipe Step Alignment, (C) Log of Similarity Metrics and Sequential Predictions, (D) History Log and Time-Causal Model, (E) Progress Tracking.}
    \label{fig:tracking}
    \Description{This is a figure that shows the pipeline of OSCAR. In the figure, it is a flow chart that show OSCAR starts from extracting cooking video to cooking images, and align cooking images with object status and recipe steps that were extracted from orginal recipe and LLM by using VLMs. It then extract the similarity metrics of alignment and save it inside a log file that stores all predicted steps. It also pass the data through a time-causal model and extract completed steps, completed steps, missing steps, and remaining steps, and save that back in the log file. It then generate the progress tracking that includes current step, completed steps, remaining steps, and missing steps.}
\end{figure*}

\section{Technical Approach: Object Status Recognition for Recipe Progress Tracking}
To explore the feasibility of using object status for recipe progress tracking, we developed OSCAR (Object Status Context Awareness for Recipes) — a technical pipeline that integrates recipe parsing, object status extraction, visual alignment, and progress prediction. While OSCAR is not a deployed assistive system, it serves as a technical probe to evaluate how object status information can enhance vision-language models for cooking progress tracking. In this section, we describe the design of OSCAR and its components.

\subsection{Recipe Formatting and Object Status Extraction}\label{Recipe Formatting and Object Status Extraction}
From prior research, we learned that recipes contain formatting differences to present recipe steps information \cite{li2024recipe}. For example, they might indicate recipe step numbers in various ways as `Step 1:', `1.' or `Step 1.' In OSCAR, We first utilized GPT-4o \cite{achiam2023gpt} to re-format the recipe steps (Figure \ref{fig:tracking}(A)). We then save the recipe steps as a variable inside OSCAR (Figure \ref{fig:tracking}(A)).

OSCAR then extracts object status information of a recipe by analyzing the ingredient list in conjunction with the associated cooking actions \cite{xue2024learning}. The extracted object statuses are represented in a structured format, typically as a combination of [verb] + [noun], such as `chopping carrots' or `sautéing mushrooms' \cite{xue2024learning}. This format allows OSCAR to clearly define the actions that are performed on each ingredient \cite{xue2024learning}. For example, a recipe that includes steps such as `whisking eggs' or `peeling potatoes' will have these actions identified and recorded to maintain a clear and actionable understanding of the object's status. We then saved these object statuses along with the recipe steps that they belong to as a variable inside a JSON file, and a single step can contain multiple object status pairs, such as `peeling carrots,' `chopping carrots,' and `storing carrots.' 

\subsection{Visual Data Extraction and Recipe Step Alignment}\label{Visual Data Extraction and Recipe Step Alignment}
After formatting recipe steps and extracting object statuses, OSCAR processes visual data by directly using the image input or extracting relevant frames from video clips using OpenCV (Figure \ref{fig:tracking}(B)) \cite{bradski2000opencv}. Once the visual content has been processed, OSCAR separately aligns recipe steps and object status with the corresponding visual frame (Figure \ref{fig:tracking}(B)). This alignment can be achieved through different Vision-Language Models (VLMs). In this paper, we applied two SOTA VLMs--CLIP (Contrastive Language-Image Pre-Training) \cite{radford2021learning} and SigLIP (Sigmoid Loss for Language Image Pre-Training) \cite{zhai2023sigmoid}--in OSCAR to show the performance differences and evaluate the approach of using object status to track recipe progress. These models are adept at associating textual descriptions with visual data, allowing OSCAR to compare the similarity scores between visual frames and the object statuses across different recipe steps. For example, when a recipe step involves `chopping tomatoes,' OSCAR uses VLMs to identify the frame that best matches this action by comparing it to other potential matches within the video. OSCAR then predicts the recipe step to which the frame corresponds based on the highest similarity score. This process ensures that visual content is accurately aligned with the progression of the recipe, reducing the chances of errors or confusion during cooking. Through the alignment, OSCAR will generate two lists of similarity scores, one for recipe steps and one for object status. OSCAR then averaged the similarity metrics to evaluate the enhancement of using the object status for the prediction of the recipe step \cite{luo2021clip4clip}.

\subsection{Log of Similarity Metrics and Sequential Predictions}
Based on the similarity metrics generated for the alignment of the recipe steps, OSCAR generates a detailed prediction log for each frame (Figure \ref{fig:tracking}(C)). The history log of each image frame is contained in a JSON file, and each entity includes the similarity scores with all steps, the predicted step info, the predicted step number, the completed steps, any missing steps, and the remaining steps based on previous predictions. The missing step is determined by having a gap between the last completed step and the current step. This log serves as a record of the cooking process, which can be used to generate a progress tracker and support context-sensitive queries (Figure \ref{fig:tracking}(C)). For example, a user might ask, 'What step am I in?' or 'Have I already sautéed the mushrooms?' OSCAR uses the prediction log to provide accurate and context-sensitive responses, guiding the user through the recipe. This feature can be beneficial in complex recipes where multiple steps or actions might overlap or occur simultaneously.

\subsection{Time-Causal Model}\label{Time-Causal Model}
After updating the log of similarity metrics and sequential predictions each time with a new image frame,
we applied a time-causal model \cite{stephan2020time,prabhakar2010temporal,pearl2018book}, to improve the temporal coherence of predictions, ensuring that OSCAR's output aligns with the natural flow of cooking tasks (Figure \ref{fig:tracking}(D)). By enforcing sequential logic, the model is designed to reduce false positives and prevent the misprediction of duplicate or out-of-order steps. Specifically, if a user completes Step N in the recipe, the model prevents the prediction of Step N-X (i.e., an earlier step in the sequence) unless that earlier step was never predicted in the first place. This ensures that the system adheres to the expected order of operations and avoids unnecessary confusion or redundant prompts for the user. OSCAR then update the progress tracking information back to the history log based on the time-causal model. This time-causal approach not only enhances the accuracy of recognizing completed steps but also strengthens the model's ability to track and predict the appropriate next action, thereby supporting a smoother and more intuitive cooking process for individuals.


\subsection{Progress Tracking}

To support reasoning about recipe state over time, OSCAR maintains a prediction log that records similarity scores between each visual frame and recipe steps. This log includes both raw similarity outputs from vision-language models and derived metadata about likely step completion, ordering, and gaps. Using this information, we apply a time-causal model to enforce temporal coherence — filtering out improbable jumps, duplicates, or regressions in step predictions based on the logical structure of recipes. From this history log, OSCAR computes a real-time snapshot of the user’s position within the recipe. Specifically, it identifies:

\begin{itemize}
    \item \textbf{Current Step:} The most likely step being performed in the current visual frame.
    \item \textbf{Completed Steps:} All steps with high prediction confidence and valid temporal alignment.
    \item \textbf{Remaining Steps:} Steps not yet observed or predicted, and appearing after the current step.
    \item \textbf{Missing Steps:} Steps that were skipped or not recognized in the expected sequence.
\end{itemize}

This structured tracking enables a time-aware understanding of progress and allows downstream applications to reason about what has happened before, what is happening now, and what comes next. An illustration of this output is shown in Figure~\ref{fig:tracking}(E).





\section{Evaluation on Large-Scale Recipe Instruction Videos}
To evaluate the feasibility and performance of OSCAR in tracking cooking progress through object status recognition, we conducted a large-scale evaluation using YouCook2, a widely used instructional cooking video dataset annotated with temporal recipe steps. This experiment aims to quantify how incorporating object status and time-causal modeling improves the alignment of recipe steps with visual frames, in comparison to baseline vision-language model (VLM) approaches.

\subsection{Dataset Description}
\subsubsection{YouCook2 Dataset}
To validate OSCAR's ability to use object statuses to predict recipe steps in visual frames, we used the YouCook2 dataset \cite{ZhXuCoAAAI18}. The YouCook2 dataset is a comprehensive collection of cooking videos, totaling 176 hours, with each video averaging 5.27 minutes long \cite{ZhXuCoAAAI18}. All the videos are untrimmed, and the length can be up to 10 minutes, they were recorded using handheld or stationary cameras \cite{ZhXuCoAAAI18}. Each video segment is temporally localized with start and end times for each procedural step. The dataset includes diverse metadata, such as video duration, the number of recipe steps per video, segment duration, and the number of words per sentence \cite{ZhXuCoAAAI18}. This diversity in content and metadata makes YouCook2 an ideal dataset to assess OSCAR’s ability to handle varied and complex cooking scenarios \cite{ZhXuCoAAAI18}. The dataset is publicly available for download at (http://youcook2.eecs.umich.edu/).

\subsubsection{Dataset Processing and Curation}
The original YouCook2 dataset did not include ingredient information in its metadata \cite{ZhXuCoAAAI18}. To address this, we manually reviewed each cooking video to extract ingredient details, sourcing information directly from the video content or the publisher’s notes on YouTube. In addition, we filtered out videos that were irrelevant to the cooking process, such as those featuring chef discussions or promotional content that are not related to cooking. After this curation process, we compiled a refined dataset consisting of 173 YouTube videos (Y1-Y173) with comprehensive ingredient lists and step-by-step instructions. These videos have an average step number of 7.7, which matches the average step of the original dataset, with minimal recipe steps of 5 and a maximum of 15. The list of videos and corresponding recipe information will be released in the camera-ready version.

\subsection{Methodology}
\subsubsection{Baseline Prediction}
Each video is annotated with timestamps of the start and end of each recipe step. For each step, we divide the video clip belonging to that step into five equal segments and randomly select a frame from each segment \cite{tong2022videomae}. We use this approach to reduce the impact of noise in the data. We applied a blur filter to choose the frame with the least blur adjacent to the selected frame in each segment \cite{bradski2000opencv}. To evaluate and understand the generalizability of how object status information can enhance the performance of VLMs, we used two SOTA VLMs in the evaluation, CLIP \cite{radford2021learning} and SigLIP \cite{zhai2023sigmoid}. As baseline methods, we then calculated the similarity score between each frame and all the recipe steps, each frame would have a list of similarity scores\cite{zhu2019r2gan}. Each time, one frame from each segment is processed, which includes five at a time for the step. The average of these similarity scores was used to determine the predicted step with the highest overall similarity \cite{zhu2019r2gan,marin2021recipe1m+}. If the predicted step aligns with the groundtruth, we will mark 100\% for this step; otherwise, we will mark 0\%. This process was repeated three times per step to ensure consistency in the results, and we used the average accuracy of the three times as the final prediction accuracy for each step \cite{zhu2019r2gan,marin2021recipe1m+}. We then repeat the same process for the rest of the recipe steps, as well as all other recipe instruction videos. Repeating the prediction process three times per step helps mitigate variability across frames within dynamic cooking scenes, where key objects may be temporarily obscured by motion or occlusion. Averaging the results across these repetitions yields a more stable accuracy estimate by reducing the influence of transient visual noise such as blur or obstruction \cite{shi2021temporal}.

\subsubsection{OSCAR with Object Status and Time-Causal Modeling}
To evaluate the usefulness of object statuses, the status information of the object was extracted from the ingredient list and the corresponding cooking steps (Section \ref{Recipe Formatting and Object Status Extraction}). Using the same video frames selected in the baseline method, we further calculated the similarity score between the frames and the object statuses using both CLIP \cite{radford2021learning} and SigLIP \cite{zhai2023sigmoid} (Section \ref{Visual Data Extraction and Recipe Step Alignment}). We then utilized the time-causal model to further enhance the performance (Section \ref{Time-Causal Model}). The process of averaging the similarity scores within the five frames per step and repetition three times is the same as the baseline method. The results were then averaged with the baseline similarity scores to obtain the final prediction accuracy to understand the usefulness of the object statuses compared to only using the recipe step text as a prediction \cite{luo2021clip4clip}.

\begin{table*}[]
\resizebox{1\columnwidth}{!}{%
\begin{tabular}{l|lllll}
VLM Model & Baseline Accuracy & Baseline Standard Deviation & OSCAR Accuracy & OSCAR Standard Deviation & $\Delta$ Accuracy (OSCAR - Baseline) \\ \hline
CLIP      & 41.7\%            & 17.5\%                      & 68.0\%         & 19.0\%                   & 26.3\%       \\
SigLIP    & 62.2\%            & 18.0\%                      & 82.8\%         & 14.7\%                   & 20.6\%       \\ \hline
\end{tabular}%
}
\caption{Evaluation results with recipe instruction videos.}
\label{table:youtubetesting}
\end{table*}

\subsection{Results}
\subsubsection{Baseline Performance}
Among 173 annotated instructional videos (Y1-Y173), the baseline results revealed an average accuracy of 41.7\% (SD = 17.5\%) for recipe step predictions using CLIP (Table \ref{table:youtubetesting}). SigLip, on the other hand, achieved an average accuracy of 62.2\% (SD = 18.0\%) (Table \ref{table:youtubetesting}). These results highlight the challenges of accurately predicting the visual data with recipe steps individually, particularly in cases where the visual content is ambiguous or lacks clear cues for specific actions. 

\subsubsection{OSCAR Performance}
By incorporating object status information and utilizing a time-causal algorithm, OSCAR outperforms the baselines with a large margin in terms of prediction accuracy. Specifically, the average accuracy for CLIP increased from 41.7\% of the baseline to 68.0\% (SD = 19.0\%), while SigLIP’s performance rose from 62.2\% to 82.8\% (SD = 14.7\%) (Table \ref{table:youtubetesting}). These improvements demonstrate the effectiveness of using object status to enhance the ability to accurately track and predict recipe steps, especially in complex cooking scenarios.

\subsection{Why Object Status Improves Step Prediction in Large-scale Recipe Instruction Videos}
\subsubsection{Disambiguating Incomplete or Occluded Visual Scenes}
By analyzing OSCAR's performance improvements compared to the baseline with 26.3\% for CLIP and 20.6\% for SigLIP (Table \ref{table:youtubetesting}), we found that the performance where OSCAR outperformed was because recipe text may reference objects or tools not directly visible in the frame. For example, some recipes instruct the use of specific tools (e.g., spatula, spoon), but these tools can be difficult to detect when obscured by other objects (e.g., soup) or when the user's hand position occludes the tool’s shape. With the use of object states, visual information can complement missing details that are not directly visible with the recipe context (e.g., diced beef in the soup). As a result, this can improve the accuracy of aligning visual frames with the corresponding recipe steps.

\subsubsection{Differentiating Visually Similar or Repetitive Actions}
Additionally, we found that implementing a time-causal model also improved performance, since some cooking steps involve similar or duplicate actions or objects at different stages of the recipe. For instance, in a stir-fry recipe for steak and vegetables, minced garlic is first added to the pan along with the steak. After a few minutes, the cook sets aside the partially cooked steak, adds more minced garlic to cook the vegetables, and then returns the steak to the pan once the vegetables are nearly done. This method ensures the steak isn't overcooked since vegetables take longer to cook. In this scenario, minced garlic is added twice, which appears visually identical. However, by incorporating a time-causal algorithm, we can distinguish between the two actions, even though they involve duplicate steps.

\begin{figure}[t]
    \centering
    \includegraphics[width=0.6\columnwidth]{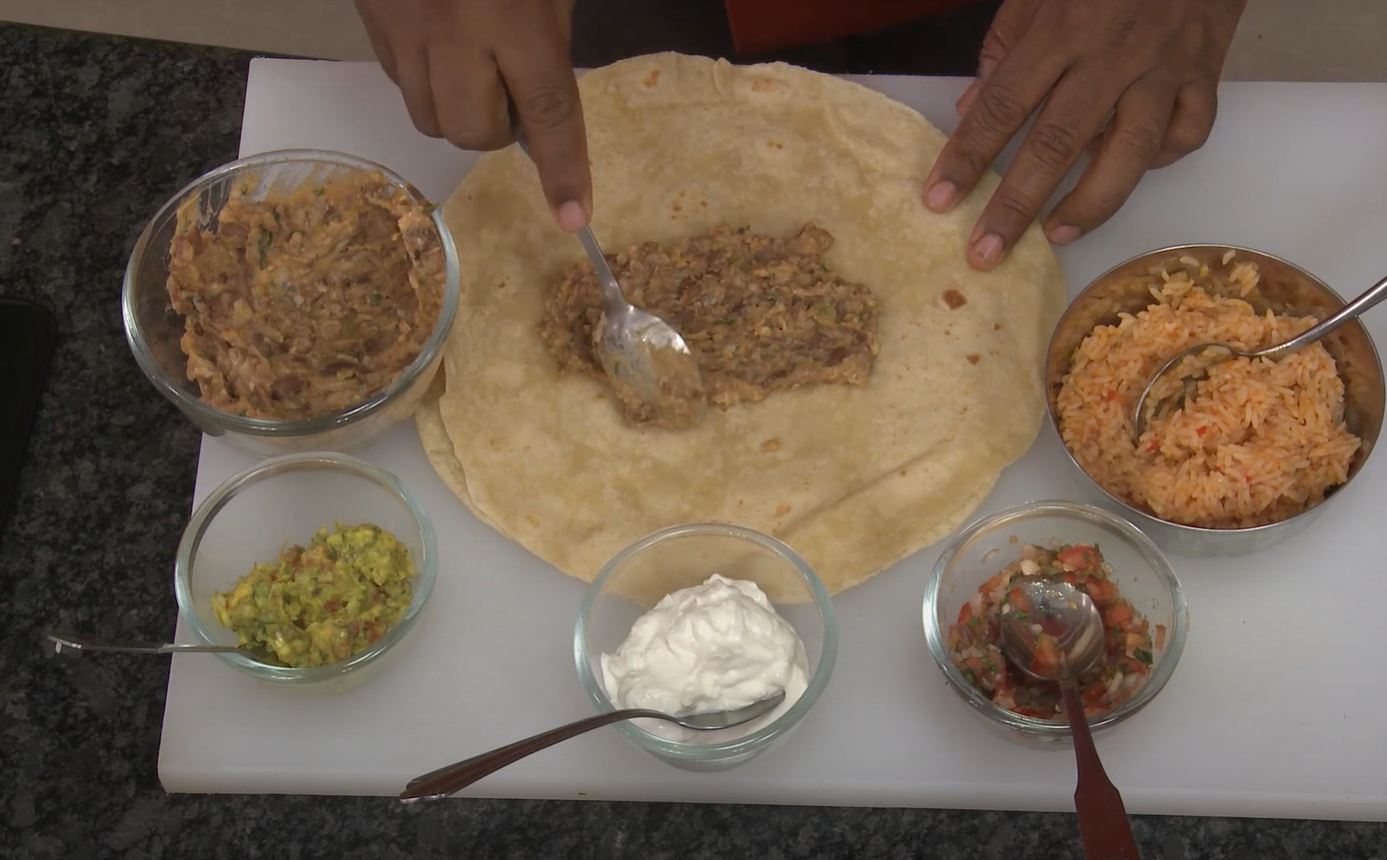}
    \caption{Image frame of Y19 that shows a top-down view with all ingredients of making the burrito that impact the prediction accuracy through object status.}
    \label{fig:burrito}
    \Description{This is an image show a person is trying to add food into the tortilla on a cutting board. There are five small bowls surround the tortilla. There are rice, fried beans, salsa, sour cream, and guacamole.}
\end{figure}

\begin{figure}[t]
    \centering
    \includegraphics[width=1\columnwidth]{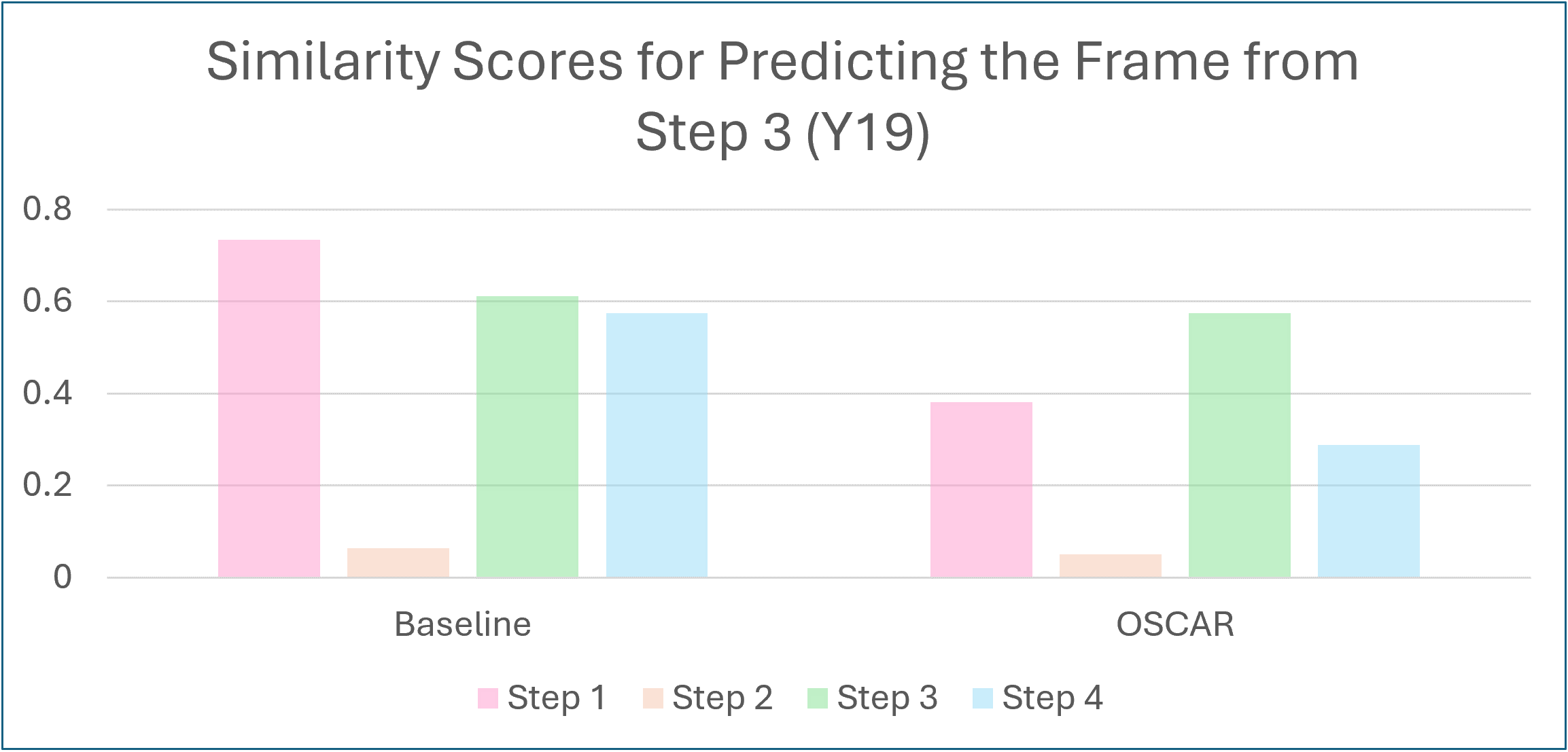}
    \caption{Similarity scores for predicting the frame from step 3 (Y19), it shows step 3 is correctly predicted by OSCAR, but not the baseline.}
    \label{fig:resultsOSCAR}
    \Description{This shows a bar chart of similarty scores for predicting the frame from step 3 (Y19). It has two groups of bars and each has four, color coded for step 1 - 4. The left is for baseline and the right is for OSCAR. The baseline one show step 1 is higher than step 3, than step 4, than step 2. The right one shows step 3 is higher than step 1, than step 4, than step 2.}
\end{figure}

\subsubsection{Disambiguating Visually Cluttered Frames with Overlapping Ingredients}

In visually cluttered scenes—such as top-down views where all ingredients appear simultaneously—baseline models often struggle to determine the correct recipe step. In Y19, where rice, refried beans, salsa, sour cream, and guacamole were visible at once (Figure~\ref{fig:burrito}), SigLIP achieved only 25\% accuracy, misclassifying most steps due to overlapping visual cues. OSCAR significantly improved performance by incorporating object status. For example, in step 3, baseline similarity scores were too close across steps (e.g., 0.734, 0.064, 0.612, 0.575), leading to misalignment. In contrast, OSCAR's object status-based scores created clearer distinctions (0.381, 0.050, 0.575, 0.288) (Figure~\ref{fig:resultsOSCAR}), enabling more accurate prediction. Even when object status alone wasn’t enough to resolve repeated steps—such as "repeat the steps for each tortilla and serve"—adding the time-causal model corrected the prediction. This case illustrates how object status provides critical disambiguating cues in cluttered frames, and how temporal modeling helps resolve step repetition.

The analysis demonstrates that both object status and time-causal model enhance the prediction performance for both CLIP and SigLIP models significantly with more than 20\% enhancement.

\subsection{Factors Affecting Prediction Performance}
While OSCAR significantly outperforms baseline models, our analysis revealed several factors that limit the effectiveness of vision-language models (VLMs) when aligning video frames with recipe steps. We conducted a detailed error analysis to identify common challenges, comparing object status descriptions against visual content across underperforming videos.

\begin{figure}[b]
    \centering
    \includegraphics[width=1\columnwidth]{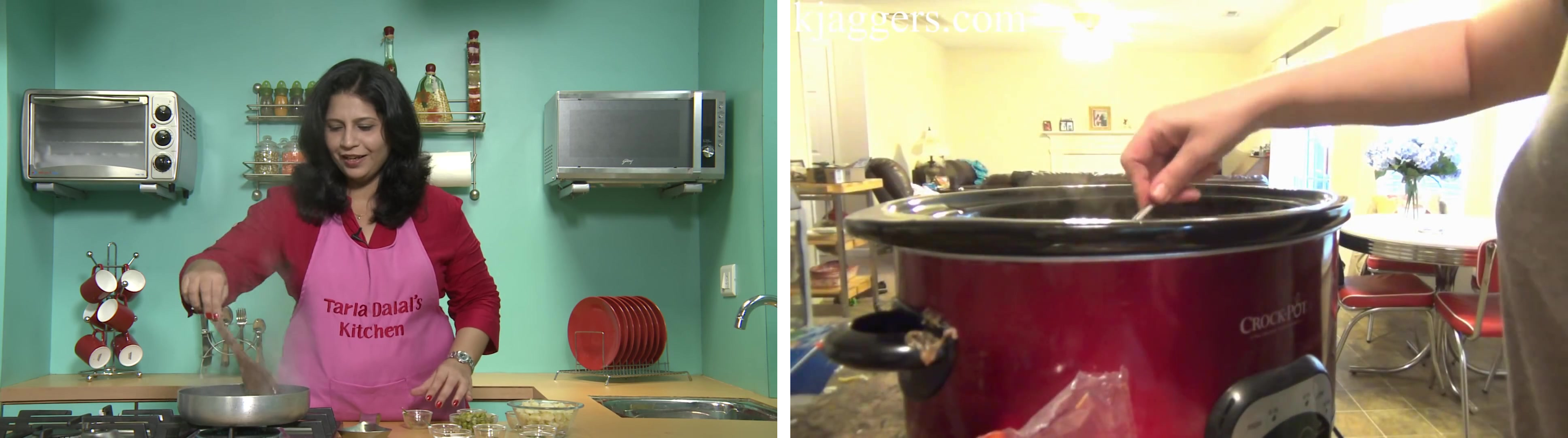}
    \caption{Image frames of cooking video that shows an exocentric view of the cooking steps. Left: This figure shows one frame in Y28, in which the exocentric view fully shows the ingredients clearly. Right: This figure shows one frame in Y33 that the person is trying to cook food in a slow cooker, and a side view cannot show what is being cooked.}
    \label{fig:exocentric}
    \Description{The image presents two different cooking scenarios. On the left, a woman wearing a pink apron that reads "Tarla Dalal's Kitchen" is seen stirring ingredients in a pan on a stovetop in a well-organized, modern kitchen. Small bowls of ingredients are placed on the counter. On the right, a close-up shot shows a person's hand stirring food in a red Crock-Pot.}
\end{figure}

\subsubsection{Exocentric vs. Egocentric View}
From the evaluation, we observed that the performance of our model using both CLIP and SigLIP dropped for some videos because certain frames in the cooking videos captured an exocentric view, which presented a broad, external perspective of the kitchen environment. For example, in Y28 (Figure \ref{fig:exocentric} left), the performance of OSCAR with CLIP was only 22.2\% compared to the average of 68.0\%. In these frames, all ingredients were displayed on the kitchen table rather than focusing closely on the specific cooking actions being performed. This wide-angle view included not only the ingredients but also the surrounding kitchen space, which made it challenging for VLMs to accurately associate these frames with the correct steps in the recipe. As a result, there shows an increase in false-positive rates, where these frames were mistakenly linked to different recipe steps.

Besides creating false positives from all ingredients, the exocentric views often resulted in cooking tools, such as pots, pans, and utensils, partially or completely obscuring the ingredients being cooked. For example, Y33 shows a performance of 55.6\% with OSCAR and SigLIP while the average was 82.8\%. This obstruction made it even more difficult for VLMs to correctly identify the status of the ingredients and the corresponding cooking actions. The lack of a clear view of what was happening in the pots or pans due to these obstructions compounded the challenge of accurately predicting the sequence of cooking steps, leading to a decrease in model performance (Figure \ref{fig:exocentric}. right).

\begin{figure}[t]
    \centering
    \includegraphics[width=1\columnwidth]{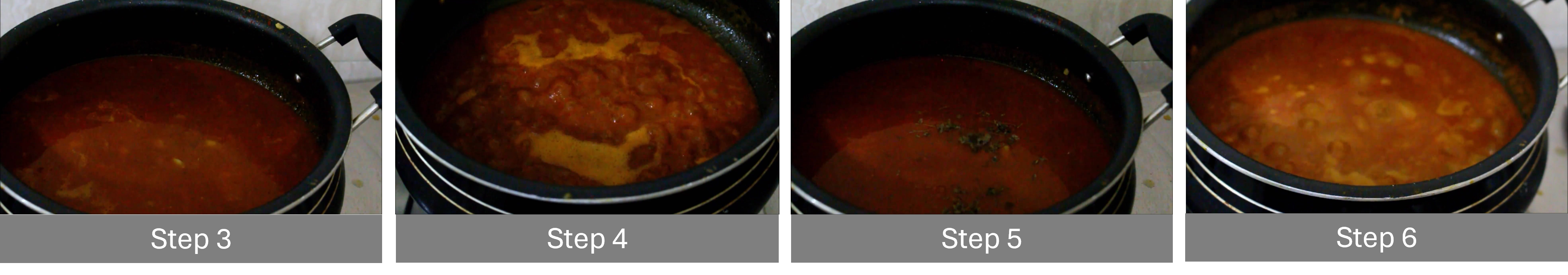}
    \caption{Image frames of cooking video that show similar visual frames of different steps that caused misprediction.}
    \label{fig:dishtype}
    \Description{This figure includes four subfigures marked from step 3 to step 6. They all show a pot is boiling soup inside. In Step 4 and 6, we can see that there might be some beans inside, and ther are some possibly herbs in step 5.}
\end{figure}

\subsubsection{Subtle Visual Changes Across Steps}
We found that the type of dish being prepared had an impact on the performance, particularly when there were only subtle visual differences between steps. For certain types of dishes, the lack of clear and distinguishable visual cues made it difficult for VLMs to identify and track the status of objects accurately. For example, in dishes such as soups (Y5), which typically involve mixing or simmering ingredients together in a pot, the changes between steps can be very gradual and subtle (Figure \ref{fig:dishtype}). It has a performance with the baseline with SigLIP of 54.2\%, which is below the average. As ingredients are mixed and cooked, the appearance of the contents in the pot often becomes homogeneous, making it difficult for VLMs to discern individual ingredients or specific object statuses (Figure \ref{fig:dishtype}). The lack of distinct, observable changes in the visual state of the ingredients leads to fewer visual cues that VLMs can use to accurately align the cooking steps with the recipe instructions (Figure \ref{fig:dishtype}).

\begin{figure}[b]
    \centering
    \includegraphics[width=0.6\columnwidth]{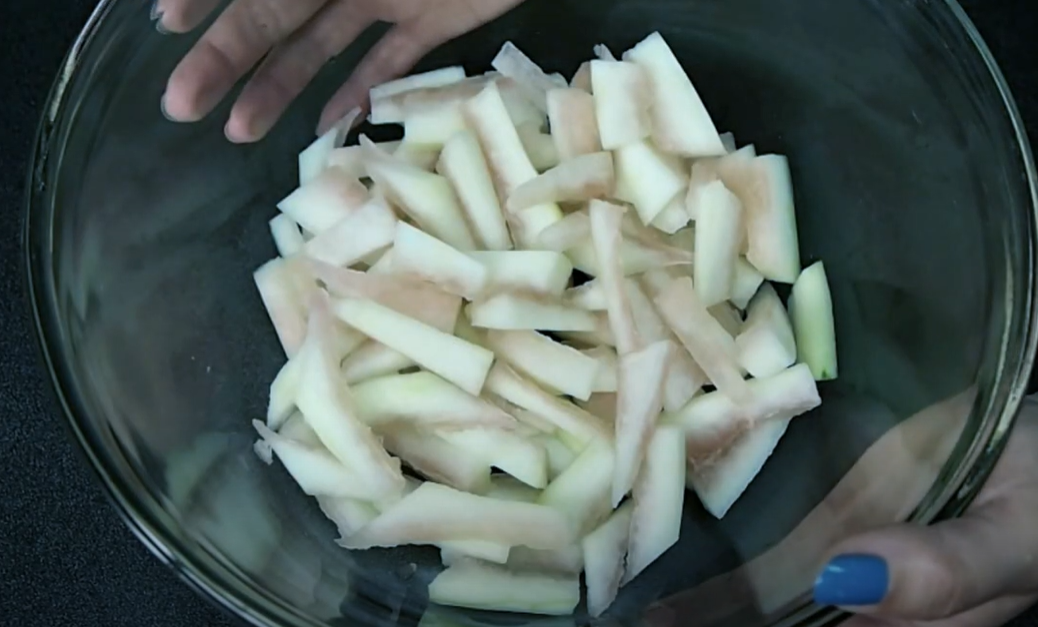}
    \caption{Image frame of cooking video that shows misprediction because of the descriptions against the norm. This figure shows the recipe of making Korean watermelon rind kimchi.}
    \label{fig:misprediction}
    \Description{This figure shows a large bowl with watermelon rind slices.}
\end{figure}

\subsubsection{Uncommon Object States or Descriptions}
Our analysis reveals that existing vision-language models are often trained with generic descriptions and standard norms, which limits their ability to accurately identify and interpret objects and their states in real-world cooking scenarios. For instance, in a Korean Watermelon Rind Kimchi recipe video (Y58), the CLIP model demonstrated limitations (Figure \ref{fig:misprediction}). It only achieved a 20\% accuracy rate through OSCAR with CLIP when identifying the watermelon rind as an ingredient, which is starkly lower compared to its average accuracy of 68\% across other, more commonly recognized ingredients. This discrepancy is likely due to the model's difficulty in recognizing the watermelon rind, which is white and visually distinct from the green watermelon skin typically associated with watermelons in most training datasets (Figure \ref{fig:misprediction}). Instead of identifying the rind based on its unique characteristics and contextual use in the recipe, the model's training on standard images led it to misidentify or overlook it. These models typically rely on large datasets that contain images of common objects and ingredients in their most typical forms \cite{nayak2024benchmarking}, focusing on standard visual characteristics. As a result, when these models are confronted with less common object states or atypical features—such as variations in color, texture, or shape—they tend to perform poorly.

\begin{figure}[t]
    \centering
    \includegraphics[width=0.6\columnwidth]{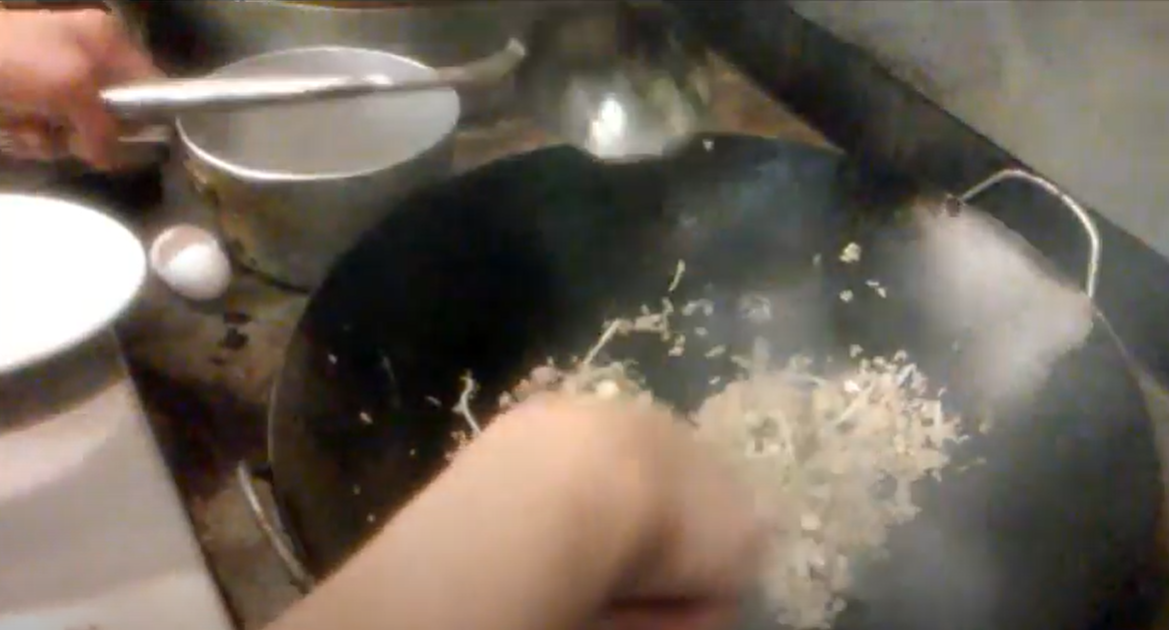}
    \caption{Image frames of cooking video that shows cooking BBQ pork fried rice and the cooking steam blurred the camera's view.}
    \label{fig:lowquality}
    \Description{This figure shows a person is trying to cook rice inside a huge wok. The person is holding a wok spatula in the left hand. It is a bit foggy on the right side of the image.}
\end{figure}

\subsubsection{Low-Quality Video and Cooking Artifacts}
The video dataset of YouCook2 included cooking recipe videos uoloaded in large span of timeframe \cite{ZhXuCoAAAI18}. In our evaluation, we found that low resolution data can impact the performance of predicting object statuses through vision-language models. For example, Figure \ref{fig:lowquality} presents one of the frames for making BBQ pork fried rice (Y48) and the video was created nine years ago and presented in 480P. In the evaluation, the performance of OSCAR with SigLIP only performed for 42.9\% while the average was 82.8\%. Furthermore, the visual impact of cooking can affect the performance of tracking recipe steps (e.g., hand occlusion, cooking steam (Figure \ref{fig:lowquality})). Figure \ref{fig:lowquality} showed the process of making BBQ pork fried rice, and it generates water vapor when the cook is stir-frying the rice, which causes the camera lens to become blurry. This results in a loss of visual clarity and detail, making it difficult to accurately align the cooking video with the recipe steps. The steam obscures important visual cues and object statuses, such as the texture and color changes of the ingredients, which are crucial for understanding the cooking process and ensuring the recipe is followed correctly.

\section{Evaluation with Real-world Non-visual Cooking Dataset}
\label{Evaluation with Real-world Non-visual Cooking Dataset}
To assess OSCAR's effectiveness beyond curated online videos, we further evaluated its performance in real-world cooking scenarios by people with vision impairments. Unlike sighted cooking practices, non-visual cooking often involves unique behaviors — such as tactile exploration, prolonged object localization, or personalized tool use — that introduce new challenges for vision-based step tracking~\cite{li2021non}. This evaluation aims to examine not only OSCAR’s prediction accuracy in natural settings but also to surface design considerations for future assistive systems that support non-visual cooking practices.

\begin{figure}[t]
    \centering
    \includegraphics[width=1\columnwidth]{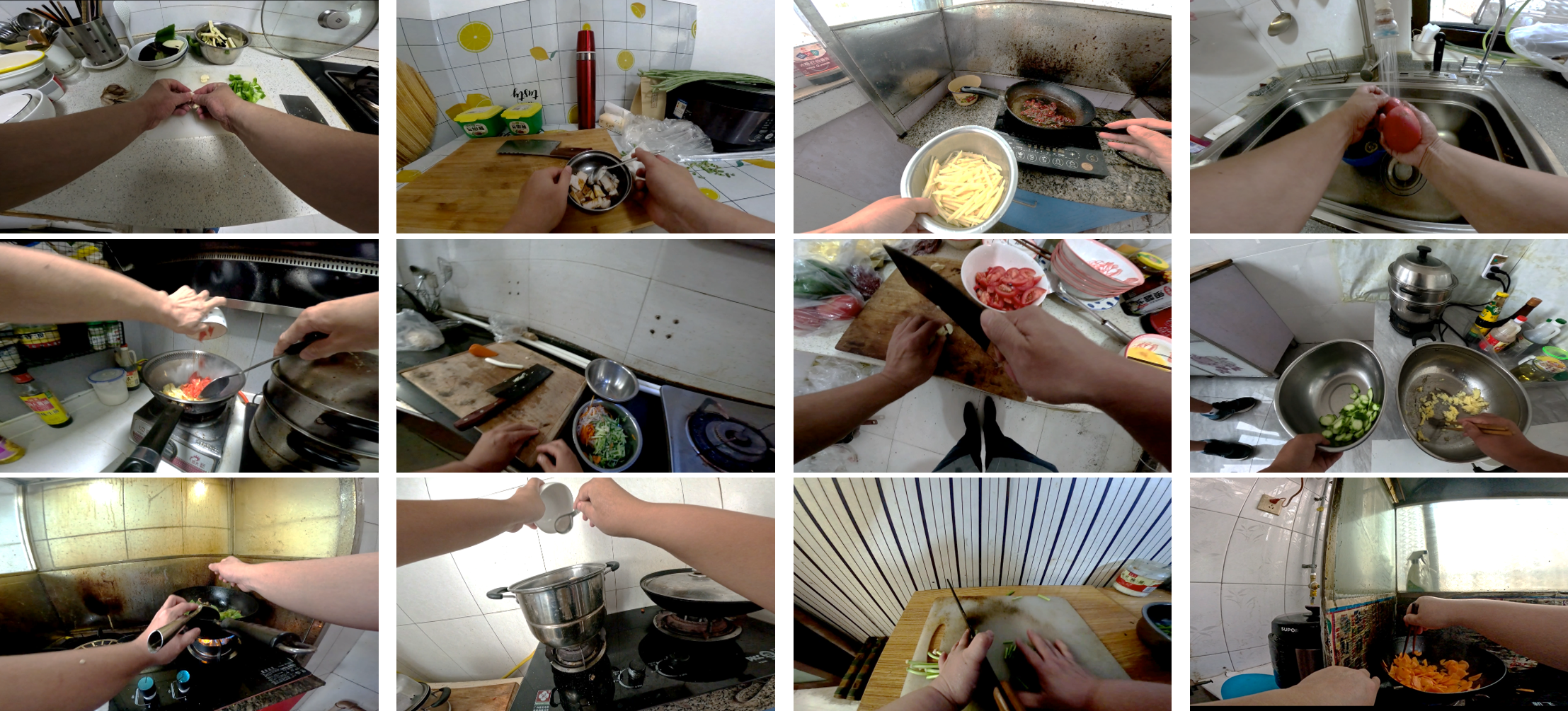}
    \caption{Thumbnail of the non-Visual cooking dataset of 12 videos by people with vision impairments.}
    \label{fig:thumbnail}
    \Description{The image displays a grid of twelve scenes (3 x 4) showing a person's hands performing various cooking tasks from a first-person perspective. These tasks include chopping vegetables, washing ingredients, stirring food in a pan, and mixing ingredients in bowls.}
\end{figure}

\subsection{Dataset Overview}
To provide the benchmark for evaluating different VLMs for non-visual cooking, we collected a new dataset of 12 cooking videos (V1–V12) featuring people with vision impairments (6 male, 6 female, average age of 43) preparing meals in their own kitchens (Figure~\ref{fig:thumbnail}). Four of them are legally blind and eight are totally blind. Each participant cooked independently, using their preferred techniques and tools to make one everyday dish. Videos ranged from 14 to 51 minutes, with an average of 8.1 recipe steps per video. Videos were recorded using a GoPro 11 chest-mounted camera in 5.2K resolution for high visual fidelity and later downsampled to 1080P for evaluation. Following the annotation protocol used in YouCook2~\cite{ZhXuCoAAAI18}, we extracted ingredients and recipe steps, then annotated the start and end timestamps for each step to support precise step alignment to evaluate OSCAR. For example, one video details washing, chopping, and assembling a salad. Participants were compensated with \$40 USD by completing the cooking session. The data collection was approved by our Institutional Review Board (IRB). Here is the dataset link: \url{https://huggingface.co/datasets/limingz3/Non-visual_Cooking_Video_Dataset}

\subsection{Methodology} 
Our evaluation followed the same experimental pipeline as the YouCook2 evaluation. We processed each video to extract object status information from ingredients and recipe steps and compared OSCAR’s performance to baseline VLM predictions.

\begin{table*}[t]
\resizebox{1\columnwidth}{!}{%
\begin{tabular}{l|lllll}
VLM Model & Baseline Accuracy & Baseline Standard Deviation & OSCAR Accuracy & OSCAR Standard Deviation & $\Delta$ Accuracy (OSCAR - Baseline) \\ \hline
CLIP      & 33.7\%            & 21.8\%                      & 58.4\%         & 16.5\%                   & 24.7\%       \\
SigLIP    & 41.9\%            & 14.0\%                      & 66.7\%         & 18.7\%                   & 24.8\%       \\ \hline
\end{tabular}%
}
\caption{Evaluation results with non-visual cooking dataset.}
\label{table:nvcheftesting}
\end{table*}

\subsubsection{Baseline Prediction}
For each annotated recipe step, we divided the corresponding video segment into five equal parts and sampled one frame per segment~\cite{tong2022videomae}. A blur filter was applied to select the clearest frame~\cite{bradski2000opencv}. Using CLIP~\cite{radford2021learning} and SigLIP~\cite{zhai2023sigmoid} as baseline models, we computed the similarity between each sampled frame and all recipe steps. The highest average similarity score across the five frames determined the predicted step. Predictions were repeated three times per step for consistency, and the final accuracy was calculated by averaging the results.

\subsubsection{OSCAR with Object Status and Time-Causal Modeling}
To evaluate the contribution of object status information, we incorporated extracted object states (Section~\ref{Recipe Formatting and Object Status Extraction}) and recalculated frame-step similarity using the same video frames as the baseline. We further applied OSCAR’s time-causal model (Section~\ref{Time-Causal Model}) to refine predictions, especially for repetitive or ambiguous actions. The final accuracy for each step was obtained by averaging the baseline similarity scores with the object status-enhanced scores, maintaining the same evaluation protocol.

\subsection{Results: Evaluating Step Prediction in Real-World Non-visual Cooking}

\subsubsection{Baseline Performance}
Baseline results revealed substantial limitations in using vision-language models (VLMs) alone to predict cooking progress in non-visual cooking contexts. On average, CLIP achieved an accuracy of only 33.7\% (SD = 21.8\%), while SigLIP reached 41.9\% (SD = 14.0\%) across the 12 annotated cooking videos (Table~\ref{table:nvcheftesting}). These figures are noticeably lower than performance on the YouCook2 dataset, emphasizing the complexity and unpredictability of real-world environments where people with vision impairments cook in ways that diverge from the assumptions encoded in conventional training datasets.

These low scores reflect multiple compounding factors: longer durations spent locating tools, non-standard tool usage, repetitive motions such as tactile verification, and variable framing and lighting due to user-generated chest-mounted camera views. Together, these introduce significant ambiguity in visual input that makes it difficult for general-purpose VLMs to determine what cooking step is currently being performed.

\subsubsection{OSCAR Performance and Improvement}
Incorporating object status representations and a time-causal prediction model, OSCAR significantly outperformed the baseline models. With OSCAR, CLIP’s average accuracy improved from 33.7\% to 58.4\% (SD = 16.5\%), and SigLIP’s from 41.9\% to 66.7\% (SD = 18.7\%) (Table~\ref{table:nvcheftesting}). This represents a performance boost of +24.7\% for CLIP and +24.8\% for SigLIP. These gains were consistent across all 12 videos, and in several cases, such as video V12, OSCAR achieved a perfect 100\% prediction accuracy—60 percentage points above the baseline.

These results not only reinforce the feasibility of using OSCAR for procedural tracking in real-world cooking tasks but also highlight its strength in adapting to user-specific routines and unconventional kitchen setups. Whereas baseline models often misinterpret lengthy or non-standard interactions, OSCAR’s attention to the current state of objects and the temporal history of actions enables it to maintain alignment with the true recipe step.

\subsection{Why Object Status Improves Step Prediction in Non-visual Cooking}

Our analysis of OSCAR’s performance improvements — achieving a +24.7\% gain for CLIP and +24.8\% for SigLIP (Table~\ref{table:nvcheftesting}) — highlights the critical role of object status recognition in supporting cooking progress tracking, particularly in real-world non-visual cooking scenarios. Unlike controlled video datasets, the non-visual cooking dataset reveals rich variations in cooking behaviors, tool use, and object interaction patterns by people with vision impairments. We identify several core reasons why object status improves prediction robustness and system adaptability.

\subsubsection{Reducing False Positives from Prolonged and Exploratory Interactions}

In non-visual cooking, people with vision impairments often engage in longer interactions with objects — feeling, exploring, or verifying their location and state through touch. These extended or repeated actions can lead to false positives when vision-language models rely only on recipe steps to infer progress, as the system may mistakenly interpret prolonged handling as task completion. By incorporating object status recognition, OSCAR provides an intermediate representation that filters out incidental or exploratory interactions unless a change in object state is detected (e.g., uncut vs. chopped tomatoes). This mechanism helps disambiguate real progress from slow or repeated exploration, improving step prediction accuracy in environments with non-linear and variable cooking workflows.

\subsubsection{Accommodating Personalized Tools and Cooking Strategies}
\label{Accommodating Personalized Tools and Cooking Strategies}

People with vision impairments frequently use personalized or adapted tools that differ from standard recipe expectations \cite{li2024recipe,li2021non}. These substitutions — such as using a butter knife instead of a spatula — often lead to mismatches between the tools referenced in the recipe text and those visible in the video frame. Recipe-step-only models struggle in these scenarios because they anchor predictions on canonical tool appearances. In contrast, object status recognition focuses on the transformation of ingredients (e.g., spread beans, washed cucumber) rather than the tool performing the action. This design flexibility enables OSCAR to generalize across different user behaviors and tool choices, supporting accessibility without requiring users to conform to pre-defined cooking procedures.

\subsubsection{Supporting Inclusive and Autonomy-preserving Design}

Tracking cooking progress through object status fosters a more inclusive interaction paradigm — one that adapts to user routines rather than forcing behavior changes. People with vision impairments often develop efficient but idiosyncratic cooking strategies shaped by their skills, preferences, and environmental constraints. OSCAR’s ability to align progress tracking with these individualized practices — without penalizing tool substitution, workspace variation, or non-standard sequencing — preserves user autonomy and supports natural cooking workflows. This is critical for designing assistive systems that respect user expertise rather than enforcing rigid procedural structures.

\subsubsection{Consistency Across Real-world and Instructional Datasets}

The performance gains of OSCAR were consistent across both the real-world non-visual cooking dataset and the large-scale YouCook2 instructional dataset. This consistency highlights that object status modeling is a transferable and generalizable approach, benefiting both controlled and naturally occurring cooking scenarios. This finding underscores the importance of moving beyond recipe text alignment alone. While recipe steps provide useful guidance, the variability in real-world cooking — especially among people with vision impairments — demands systems that can reason about the physical state of the environment. Object status, combined with time-causal modeling, provides a scalable foundation for robust cooking progress prediction in diverse settings.

\subsection{Factors Affecting Prediction Performance and Design Implications}
\label{Factors Impacted the Performance and Design Implications}
While OSCAR demonstrated substantial improvements over baseline models, our evaluation on the real-world non-visual cooking dataset revealed several unique factors that challenged the performance of vision-language models (VLMs) in non-visual cooking settings. These findings offer critical insights for the design of future vision-based assistive systems.

\begin{figure}[t]
    \centering
    \includegraphics[width=1\columnwidth]{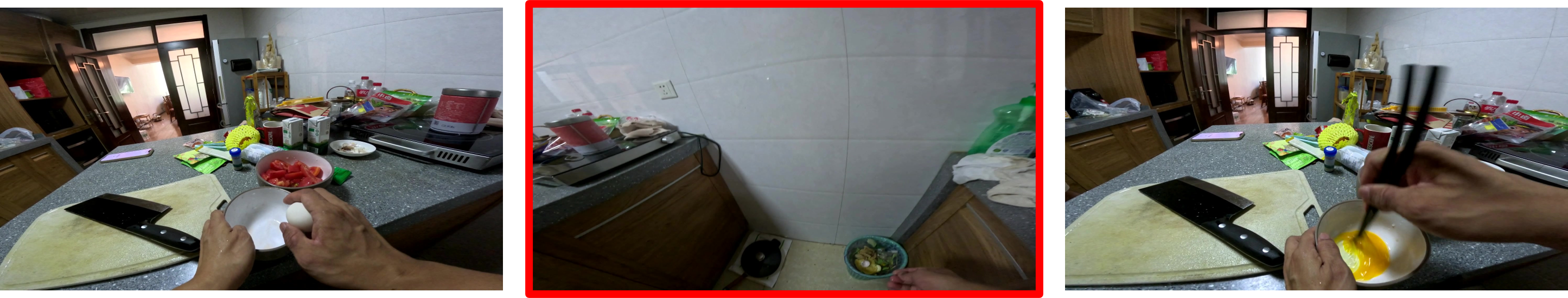}
    \caption{These three frames were captured for V4 during step 1: `Crack an egg and scramble it.' The middle frame showed the data where the blind cook was looking for a garbage bin to throw the egg shell, which got captured and impacted the performance of predicting the step.}
    \label{fig:lookingforobjects}
    \Description{The image presents a sequence of three first-person cooking scenes. In the first scene, the individual is handling a bowl of chopped tomatoes on a cutting board, with a cleaver resting nearby on a cluttered kitchen countertop, while the living room is visible in the background. In the second scene, the person appears to be discarding food waste, holding a small bowl near a garbage bin, with various cooking items and packaging visible on the counter. In the third scene, the individual is whisking eggs with chopsticks in a small bowl, continuing the cooking process, with the same kitchen countertop and supplies in the background.}
\end{figure}

\subsubsection{Implicit and Preparatory Tasks Beyond Recipe Instructions}

In real-world cooking, people with vision impairments often perform implicit or preparatory tasks that are not explicitly described in recipe steps — such as cleaning, organizing tools, or discarding waste. These actions are essential for maintaining safety, hygiene, and workflow readiness, yet they create gaps between recipe expectations and real-world behavior. For example, in V4, the participant paused during the step \textit{"crack an egg and scramble it"} to search for a garbage bin and dispose of the eggshell (Figure~\ref{fig:lookingforobjects}). Since this action was not annotated as part of the recipe step, OSCAR mispredicted the progress as 0\%. These moments introduce ambiguity in vision-based step tracking, particularly when object interaction occurs without a corresponding transformation in object status.

\textbf{Design Consideration \#1:}  The presence of frequent implicit and preparatory tasks in real-world cooking—such as cleaning, organizing tools, or discarding waste—suggests that assistive systems should be designed to recognize and accommodate these actions rather than treat them as noise. Distinguishing between core procedural steps and incidental yet essential behaviors can help models interpret natural deviations from scripted instructions without misclassifying user intent. Such consideration is critical for ensuring that systems remain aligned with the practical, adaptive workflows common in non-visual cooking.

\subsubsection{Rechecking and Securing Tools and Ingredients}

A notable pattern in the non-visual cooking dataset was participants’ frequent rechecking of ingredients and tools — a behavior deeply rooted in strategies for ensuring safety, spatial awareness, and task confidence. Unlike sighted cooking routines, where visual scanning often suffices, people with vision impairments rely on tactile feedback to confirm object identity, location, and readiness. This leads to a recurring process of touching, repositioning, or verifying objects before proceeding. For example, in video V4, after the participant chopped tomatoes and placed them in a bowl, they revisited the bowl multiple times — gently touching the tomatoes and confirming their position on the countertop — before continuing to the next step. These interactions were not part of the functional transformation of the tomatoes, yet from the model’s perspective, the repeated contact resembled the type of object manipulation seen in active steps.

These exploratory or confirming actions often occurred not only immediately after processing ingredients but also as part of users’ adaptive strategies to stay prepared and oriented for future steps. For instance, some participants proactively ensured that utensils like spoons or knives were in the expected place before they were needed. While such behaviors helped users maintain workflow continuity, vision-language models frequently misclassified these moments as step transitions, resulting in inflated or inaccurate step completion predictions.

\textbf{Design Consideration \#2:}  The frequent use of exploratory and safety-checking behaviors—such as rechecking object positions or preparing tools in advance—indicates that assistive systems should distinguish these non-transformative actions from actual step progression. When users interact with objects without changing their state, such as lightly touching or repositioning already-prepared ingredients, these moments should not be treated as evidence of step completion. Instead, systems need to integrate object status recognition with temporal reasoning and patterns of interaction that reflect common non-visual strategies. This approach can help reduce false positives in step tracking, improve alignment with user intent, and make the system more resilient to the natural variability in non-visual cooking practices.

\subsubsection{Lighting Conditions and Visibility Challenges}

Non-visual cooking environments often involve variable and uncontrolled lighting, which can significantly degrade vision model performance. For instance, in V6, suboptimal lighting reduced OSCAR’s accuracy to 40\%, well below the 66.7\% average. Two lighting issues emerged frequently: (1) strong backlighting caused models to fixate on irrelevant background features, and (2) dim environments led to poor contrast, making ingredients hard to distinguish from shadows. These conditions made it difficult for VLMs to detect object transitions or ingredient changes, undermining step prediction.

\textbf{Design Consideration \#3:} The frequent impact of poor or uneven lighting on model performance highlights the need for systems to account for real-world visibility constraints. Inconsistent lighting—such as strong backlighting or dim conditions—can obscure ingredient changes and lead to misclassification. These findings point to the importance of incorporating lighting-aware feedback mechanisms and designing for robustness under suboptimal visual conditions. Strategies such as real-time lighting quality checks, user guidance for repositioning light sources or the camera, and multimodal sensing (e.g., audio or haptics) may help ensure more consistent and reliable step tracking in varied kitchen environments.

\begin{figure}[t]
    \centering
    \includegraphics[width=0.8\columnwidth]{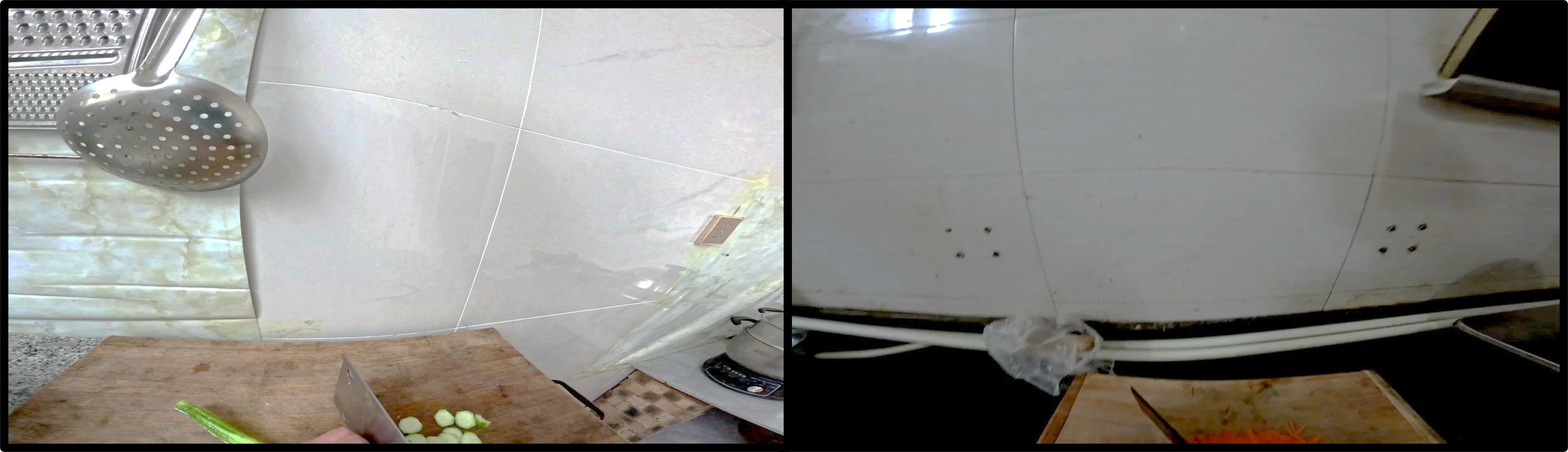}
    \caption{Illustration of objects are not captured in the center of the field of view. Left: V8 is showing `Cut the cucumber into pieces and store them in a bowl.' Right: V6 is showing `Wash the enoki mushroom, carrot, and cucumber. Shred the carrot and cucumber.'}
    \label{fig:fieldofview}
    \Description{The image shows a split view of a kitchen setting from a first-person perspective. On the left side, a metal slotted spoon hangs on the wall above a wooden cutting board, where someone is preparing vegetables, with a partially sliced cucumber visible. The tiled wall behind the cutting board shows some signs of wear and kitchen use, such as stains or discoloration. On the right side, the view shows a section of the same tiled kitchen wall, with some holes visible where something may have been previously mounted. Below the wall, pipes and plastic bags are visible, suggesting this area may be near the sink or stove. A small portion of a cutting board with food, possibly shredded carrots, is also visible in the bottom right corner.}
\end{figure}

\subsubsection{Impact of Camera Angles and Field of View} 
Inconsistent camera angles and limited field of view significantly impaired prediction accuracy in real-world non-visual cooking videos. In several cases, the camera failed to center on the active work area, leading to partial or missing views of ingredients and tools. For example, in V6, the camera was angled too high during the step \textit{“Wash the enoki mushroom, carrot, and cucumber. Shred the carrot and cucumber”}, resulting in only a partial view of the cutting board (Figure~\ref{fig:fieldofview}). This contributed to V6 showing the lowest accuracy in the dataset: 40\% with OSCAR and just 16.7\% with the baseline model. For users with vision impairments, adjusting the camera angle or position during cooking is often impractical, especially when hands are occupied. Misaligned views can distort objects or exclude them entirely, reducing the model’s ability to detect status changes and step transitions.

\textbf{Design Consideration \#4:} The frequent occurrence of misaligned or incomplete camera views underscores the need to design for hands-free, robust framing support. When the field of view fails to capture the active workspace, vision models struggle to detect object status changes and task progression, leading to substantial drops in prediction accuracy. These challenges highlight the importance of incorporating mechanisms that either assist users in maintaining appropriate framing or compensate for imperfect views—such as real-time framing feedback, multi-camera setups, or adaptive cropping techniques. Such support is especially critical in non-visual contexts, where users cannot easily verify what the camera is seeing during dynamic, hands-busy tasks like cooking.

\begin{figure}[t]
    \centering
    \includegraphics[width=0.6\columnwidth]{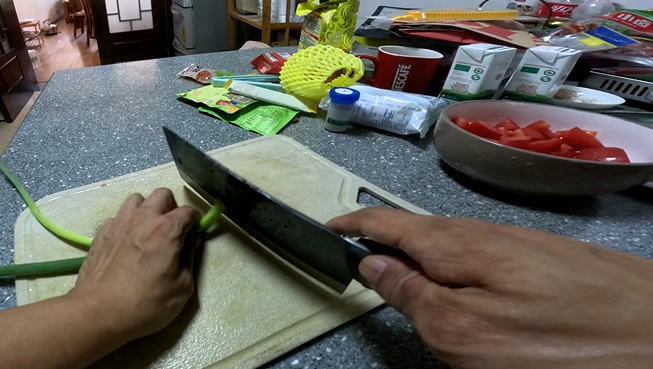}
    \caption{Illustration of multiple objects was stored before cooking.}
    \label{fig:multipleobjects}
    \Description{This figure shows the person is trying to cut a stalk of green onion. There is a bowl of chopped tomatoes on the side.}
\end{figure}

\subsubsection{Challenges with Recognizing Pre-prepared Ingredients}
People with vision impairments often adopt efficient cooking strategies, such as preparing and storing ingredients ahead of time in separate containers. While this helps streamline their workflow and improve task flow, it introduces ambiguity for vision-based systems like OSCAR, which rely on visual cues to align recipe steps with observed actions. For instance, in V4, the cook followed two steps in sequence: \textit{"Cut the tomatoes into pieces and store them in a bowl"} and \textit{"Take a stalk of green onion and cut it into pieces"}. However, during the second step, the bowl of pre-cut tomatoes remained visible in the frame (Figure~\ref{fig:multipleobjects}). This led OSCAR to misclassify the current step as a continuation of the earlier one, as the presence of tomatoes was interpreted as an ongoing action rather than a previously completed one.

This overlap between past and current ingredients illustrates the need for more sophisticated context modeling. Simply detecting object presence or status is insufficient—models must also reason about temporal sequencing and distinguish between active manipulation and background presence.

\textbf{Design Consideration \#5:} The presence of pre-prepared ingredients in the visual frame—even when they are no longer being manipulated—reveals a critical gap in how current systems interpret context. Users’ strategies, such as batch-preparing ingredients and keeping them nearby for later use, result in scenes where previously completed steps remain visually present. This makes it difficult for models to distinguish between ongoing actions and residual context. These observations point to the need for assistive systems to move beyond static object detection and incorporate reasoning about temporal context and interaction history. By understanding when ingredients have been processed and set aside, systems can more accurately interpret which actions are in progress, improving the fidelity of step alignment in real-world cooking scenarios.

\subsubsection{Summary}
Taken together, these findings highlight the unique complexities of supporting real-world non-visual cooking through vision-language models. Unlike instructional videos, non-visual cooking involves adaptive behaviors—such as exploratory touching, pre-preparation, and implicit tasks—that do not always align neatly with recipe instructions or visual cues. Performance was also sensitive to environmental variables like lighting, camera positioning, and field of view, which are difficult for users with vision impairments to control. These insights underscore the importance of designing assistive systems that go beyond frame-by-frame recognition: systems must model temporal context, support user autonomy, and adapt to the idiosyncrasies of real-world practice. By integrating object status tracking with reasoning about intent, memory, and environment, future systems can better align with the actual workflows and needs of blind and visually impaired cooks.

\section{Discussion}
This work presents OSCAR, a technical pipeline that leverages object status recognition to support progress tracking in non-visual cooking. Our findings offer broader implications for assistive AI — particularly around modeling real-world variability, supporting user autonomy, and designing systems that adapt to the embodied and procedural nature of non-visual tasks.

\subsection{Rethinking Progress Tracking in Real-World Non-Visual Activities}
Our findings challenge the dominant paradigm of recipe progress tracking that assumes clear task boundaries, discrete action-object mappings, and linear step completion. While these assumptions hold in curated instructional videos (e.g., YouCook2), they break down in real-world non-visual cooking, where people with vision impairments often engage in exploratory, repetitive, and non-linear behaviors (Section \ref{Factors Impacted the Performance and Design Implications}). For example, participants frequently performed extended tactile searching, rechecking ingredient locations, or organizing tools for future use (Section \ref{Factors Impacted the Performance and Design Implications}), none of which directly transformed object status but were essential for task success and safety. These behaviors led to consistent mispredictions in baseline models that relied solely on aligning static recipe steps to video frames (Table~\ref{table:nvcheftesting}). This finding echoes prior accessibility research that emphasizes the non-linear, adaptive nature of non-visual activities \cite{li2024recipe,li2021non,li2023understanding,li2022feels,kianpisheh2019face}, where progress is often embodied, situational, and opportunistic rather than strictly sequential.

To support real-world assistive applications, future systems must move beyond rigid step-alignment and instead adopt dynamic progress inference models that accommodate the fluidity of non-visual workflows. This requires combining object status recognition with temporal reasoning (Section \ref{Time-Causal Model}) and possibly integrating interaction modeling (e.g., recognizing exploratory touch vs. transformative actions). Such an approach would allow systems to reason about "task readiness" or "step proximity" rather than enforcing binary step completion. Ultimately, rethinking progress tracking in this way aligns with inclusive design principles that respect user autonomy, adapt to diverse routines, and support the messy realities of everyday non-visual activities.

\subsection{Vision-Language Models in Messy Real-World Conditions}
Our evaluation revealed a stark contrast between OSCAR’s performance on curated instructional data (YouCook2) and its performance in real-world non-visual cooking settings (Section~\ref{Evaluation with Real-world Non-visual Cooking Dataset}). While vision-language models (VLMs) like CLIP and SigLIP achieved recipe step prediction accuracies above 60\% in the controlled, well-lit videos of YouCook2 (Table~\ref{table:youtubetesting}), their baseline performance dropped significantly — to 33.7\% and 41.9\% respectively — in user-generated videos recorded by blind and low vision (BLV) individuals in their own kitchens (Table~\ref{table:nvcheftesting}).

This performance gap exposes key fragilities in current VLMs when deployed outside of clean, curated datasets. In our real-world dataset, videos featured challenging visual conditions: inconsistent or dim lighting, cluttered backgrounds, and occluded ingredients or tools due to variable camera angles. More critically, BLV users engaged in adaptive behaviors — such as tactile searching, using alternative tools, or pre-preparing ingredients — that led to object states or visual frames not typically encountered in training corpora. As a result, models often misclassified steps or failed to recognize when progress had occurred.

These findings underscore a core challenge in accessibility AI: models trained on polished, idealized data rarely generalize to the messy, situated realities of lived experience. As Theodorou et al. argue \cite{theodorou2021disability}, disability-centered technologies must be evaluated and optimized for the contexts in which they are used — not just the environments in which they are easiest to test. To build truly inclusive and dependable systems, VLM research must treat robustness under user-authored, imperfect, and unpredictable conditions as a core design goal — not a downstream deployment concern. Models must be built with the variability of real-world usage in mind, especially when intended to support autonomy in daily life.

\subsection{Object Status as a Universal Design Primitive}
Our findings highlight object status recognition as a transferable design primitive for assistive technologies supporting non-visual activities. In our non-visual cooking evaluation (Section~\ref{Accommodating Personalized Tools and Cooking Strategies}), participants frequently substituted tools — using fingers to spread beans (V3) or butter knives to lift food (V5) — which led to baseline model errors (Table~\ref{table:nvcheftesting}) that assumed fixed action-tool mappings and canonical tool usage. These errors expose a limitation in systems that treat progress as a function of specific object-action pairs rather than the outcome of a transformation.

In contrast, recognizing the evolving state of objects — such as ingredients becoming chopped, bowls becoming filled, or pans becoming empty — enabled OSCAR to track progress independent of the tools or gestures used. By grounding progress in the status of objects, rather than how the status was achieved, OSCAR offers greater flexibility and better accommodates diverse user routines. This abstraction aligns with inclusive design principles \cite{li2024recipe,li2021non}, supporting user autonomy, customization, and adaptation — critical values in accessibility technology.

Importantly, object status recognition is not limited to cooking. Tasks like makeup application, cleaning, crafting, or home repair similarly involve the transformation of materials and tools through personalized strategies \cite{li2022feels,huh2025vid2coach,li2022freedom}. In these domains, users may mix techniques, skip steps, or use improvised tools, yet the meaningful unit of progress remains the changing state of the task environment. Designing assistive systems around object status — rather than predefined actions or tools — offers a more resilient and generalizable foundation for context-aware support across varied non-visual, hands-on tasks.

\subsection{Modeling Context, Memory, and Temporal Continuity}
While our real-world evaluation exposed limitations of VLMs in non-visual cooking (Section~\ref{Evaluation with Real-world Non-visual Cooking Dataset}), these issues stem not only from data quality but from fundamental modeling constraints. Current VLMs rely heavily on per-frame classification, lacking mechanisms to reason about what has already occurred or how scenes evolve over time — a shortcoming echoed in recent work on temporal grounding and task tracking \cite{xue2024learning}. For example, in V4, previously prepared ingredients (e.g., chopped tomatoes) remained in view during later steps, causing baseline models to misclassify the user’s current action. Without temporal memory or an understanding of scene history, these systems cannot distinguish between residual context and active manipulation. To better support non-visual, multi-step tasks with high ambiguity, future systems must move beyond isolated image-text alignment. Techniques such as short-term memory buffers, lightweight scene graphs, or temporal event segmentation could help systems maintain continuity, disambiguate repeated actions, and better track progress through non-linear or interleaved workflows.

\subsection{Toward Context-Aware Interaction with LLMs}
While OSCAR’s primary contribution lies in its object status-based progress tracking, our architecture also opens up promising opportunities for context-aware, conversational interaction powered by large language models (LLMs). By maintaining a structured history log—including step predictions, visual context, and object transformations over time—OSCAR can dynamically construct prompts that reflect the current cooking state. This enables LLMs to reason not only about static recipe text but also about the user's actual progress through the task. For instance, users could ask situational questions such as “What step am I on?” or “Did I already add the garlic?” and receive responses grounded in their cooking history rather than generic recipe flow. This integration represents a shift from linear recipe readers toward procedurally aware dialogue systems, where interactions are informed by both real-time observations and past actions. As LLMs continue to improve in temporal reasoning and contextual understanding, such a pipeline could support richer, personalized guidance—especially for users who rely on non-visual cues to navigate complex tasks. Future work may explore the design space of such interactions, including how to present uncertainty, verify state interpretations, and align system responses with user mental models.

\subsection{Designing for Autonomy over Standardization}
Our findings underscore that non-visual cooking is shaped not by deviation from norms, but by adaptation grounded in embodied expertise and personal strategies (Section~\ref{Accommodating Personalized Tools and Cooking Strategies}). Participants frequently modified workflows — substituting tools, reordering steps, or combining actions — not out of error, but to optimize for safety, efficiency, or tactile feedback. These adaptations often confounded baseline models that assumed fixed action-tool mappings or linear task flows. In contrast, OSCAR’s object-status-driven tracking — focusing on the evolving state of objects rather than prescriptive actions — was more resilient to user-specific variations. This finding aligns with broader accessibility research cautioning against assistive technologies that enforce normative or idealized procedures \cite{li2024recipe, li2021non}. Future systems should foreground adaptability and respect user autonomy, allowing people to work with their own tools, sequences, and strategies — rather than requiring conformance to model assumptions.

\subsection{Interaction and Camera Design for Blind Cooks}
A recurring challenge in our real-world evaluation was camera misalignment — with chest-mounted cameras often angled too high, too low, or failing to capture the primary workspace. In V6, for example, the camera missed the cutting board entirely during ingredient preparation, contributing to a low prediction accuracy of 40\% despite adequate lighting. For BLV cooks, this issue is not simply about camera placement — it reflects a fundamental interaction gap. Without visual feedback, users cannot easily verify what the camera captures during dynamic tasks like cooking. This challenge echoes prior research in blind photography, where real-time framing support — such as haptic cues, audio tones, or spoken guidance — has been shown to help users position cameras effectively \cite{jayant2011supporting, adams2014blind}. For systems like OSCAR to operate reliably in deployment, camera-aware interaction must become a core design consideration. Future systems should provide real-time feedback to guide framing, suggest re-alignment when needed, or explore multi-camera setups to increase robustness — all while minimizing user effort during hands-busy tasks like cooking.

\subsection{Expanding Accessibility Datasets for Inclusive AI}
We will release our 12-video non-visual cooking dataset to support evaluating procedural AI in real-world accessibility contexts (Section~\ref{Evaluation with Real-world Non-visual Cooking Dataset}). To more fully validate systems like OSCAR, future datasets must reflect a broader range of kitchen environments, cultural cuisines, and non-visual cooking practices. As prior datasets like ORBIT \cite{massiceti2021orbit}, the true value of such resources lies in their diversity and realism. Future efforts should broaden this foundation by capturing data across varied kitchen setups, lighting conditions, cultural cooking practices, and user expertise levels. In particular, collecting multi-session recordings from the same participants could support modeling of longitudinal learning, routine formation, and adaptation over time. Echoing principles from disability-centered dataset research \cite{theodorou2021disability}, inclusive AI development requires datasets that reflect the messiness, variability, and situated practices of real users — not idealized or sanitized tasks. Expanding datasets along these dimensions is critical for advancing robust, user-centered assistive technologies.

\section{Limitation and Future Work}
While this paper demonstrates the feasibility and performance of OSCAR for recipe tracking using object status recognition, several limitations remain. Our current evaluation was conducted entirely on pre-recorded video dataset, without direct user interaction with the system. As a result, we were not able to observe how blind users might engage with OSCAR in real-time — particularly how they interpret contextual feedback, interact with step tracking, or formulate conversational queries during the cooking process. Future work should focus on developing an accessible mobile prototype and conducting in-situ studies in users’ own kitchens. These studies will examine how users naturally interact with the system, including their input strategies, preferences for conversational support, and ways of integrating real-time feedback into non-visual workflows. Insights from these investigations will help refine OSCAR’s interactive capabilities and guide the design of context-aware assistive systems grounded in real-world use. Finally, our evaluation focused on two state-of-the-art VLMs (CLIP and SigLIP). As multimodal models continue to advance (e.g., GPT-4o), evaluating OSCAR with a wider range of models will help uncover its broader applicability. In particular, future work should explore what levels of model performance are “good enough” for practical use and define benchmarks that center user experience, not just accuracy.

\section{Conclusion}
This paper presented OSCAR (Object Status Context Awareness for Recipes), a technical pipeline that explores object status recognition as a foundation for context-aware cooking assistance for people with vision impairments. By integrating recipe parsing, object status extraction, vision-language alignment, and time-causal modeling, OSCAR enables real-time recipe progress tracking that adapts to diverse user practices and non-visual workflows. Through large-scale evaluation on instructional videos and a real-world non-visual cooking dataset recorded by blind and low vision individuals in their homes, we demonstrated that object status recognition consistently improves step prediction accuracy over baseline models, even in challenging, uncontrolled environments. Our findings highlight the limitations of current vision-language models in real-world accessibility contexts and underscore the importance of modeling temporal continuity, user autonomy, and task adaptability. We contribute both a technical pipeline for object-status-driven progress tracking and a real-world cooking dataset to support future research for non-visual cooks. Looking forward, we envision expanding OSCAR into interactive, user-facing prototypes and conducting in-situ studies to explore how BLV cooks engage with real-time feedback and context-aware guidance. Our work provides design insights for building robust, inclusive assistive technologies that align with the complexities of non-visual everyday practices.

\bibliographystyle{ACM-Reference-Format}
\bibliography{main}


\end{document}